\documentclass[runningheads]{llncs}

\usepackage{eccv}

\usepackage{lipsum}
\usepackage{xcolor,colortbl}
\usepackage{multirow}
\usepackage{array}

\usepackage{eccvabbrv}

\usepackage{graphicx}
\usepackage{booktabs}

\usepackage{marvosym}

\usepackage[accsupp]{axessibility}  

\usepackage{hyperref}

\usepackage{orcidlink}

\title{Skeleton-based Group Activity Recognition via Spatial-Temporal Panoramic Graph}

\titlerunning{Skeleton-based GAR via Panoramic Graph}

\author{Zhengcen Li\inst{1,2}\orcidlink{0000-0001-9736-7375} \and
Xinle Chang\inst{1} \and
Yueran Li\inst{1} \and
Jingyong Su\inst{1,2}\textsuperscript{(\Letter)}\orcidlink{0000-0003-3216-7027}
} 

\authorrunning{Li et al.}

\institute{Harbin Institute of Technology, Shenzhen, 518055, China \and
Pengcheng Laboratory, Shenzhen, 518055, China \\
\email{\{lizhengcen, changxinle, liyueran\}@stu.hit.edu.cn} \\
\email{sujingyong@hit.edu.cn}
}

\begin{document}
\maketitle
\begin{abstract}

Group Activity Recognition aims to understand collective activities from videos. 
Existing solutions primarily rely on the RGB modality, which encounters challenges such as background variations, occlusions, motion blurs, and significant computational overhead. 
Meanwhile, current keypoint-based methods offer a lightweight and informative representation of human motions but necessitate accurate individual annotations and specialized interaction reasoning modules. 
To address these limitations, we design a panoramic graph that incorporates multi-person skeletons and objects to encapsulate group activity, offering an effective alternative to RGB video. 
This panoramic graph enables Graph Convolutional Network (GCN) to unify intra-person, inter-person, and person-object interactive modeling through spatial-temporal graph convolutions. 
In practice, we develop a novel pipeline that extracts skeleton coordinates using pose estimation and tracking algorithms and employ Multi-person Panoramic GCN (MP-GCN) to predict group activities.
Extensive experiments on Volleyball and NBA datasets demonstrate that the MP-GCN achieves state-of-the-art performance in both accuracy and efficiency. 
Notably, our method outperforms RGB-based approaches by using only estimated 2D keypoints as input. 
Code is available at \href{https://github.com/mgiant/MP-GCN}{https://github.com/mgiant/MP-GCN}.




\end{abstract}    
\section{Introduction}
\label{sec:intro}

    Group Activity Recognition (GAR) is an essential task in video understanding and is widely applied in surveillance, social scene understanding, and sports analysis~\cite{choi2009what}. The objective of GAR is to classify the collective activity of a group of actors from a given video clip. 
    This task is challenging because an optimal solution requires not only the localization of key roles within a complex scene, but also the effective modeling of spatial-temporal contextual information~\cite{ibrahim2016volleyball}.

    The majority of preceding methods for GAR~\cite{wu2019learning,gavrilyuk2020actortransformers,yuan2021learning,kim2022detectorfree,han2022dualai} are based on the RGB modality or the fusion of RGB and other modalities. These methods typically employ a CNN backbone to extract individual features with human bounding boxes and model interactions between actors using RNNs, GNNs, or Transformers.
    Recognizing actions from RGB data can be challenging due to background variations, motion blurs and occlusions~\cite{sun2023surveyHARmultimodal}. Furthermore, RGB videos often contain redundant information and are large in data size. 
    In group activity scenarios, high resolution and frame rate are crucial for accurate detection and recognition. However, most RGB-based methods have to reduce the batch size~\cite{yuan2021spatiotemporal} or employ frame sampling~\cite{han2022dualai,kim2022detectorfree} to manage the high computational and memory demands. These compromises significantly constrain the application of RGB-based GAR methods.



\begin{figure}[t]
    \centering
    \includegraphics[width=0.85\linewidth]{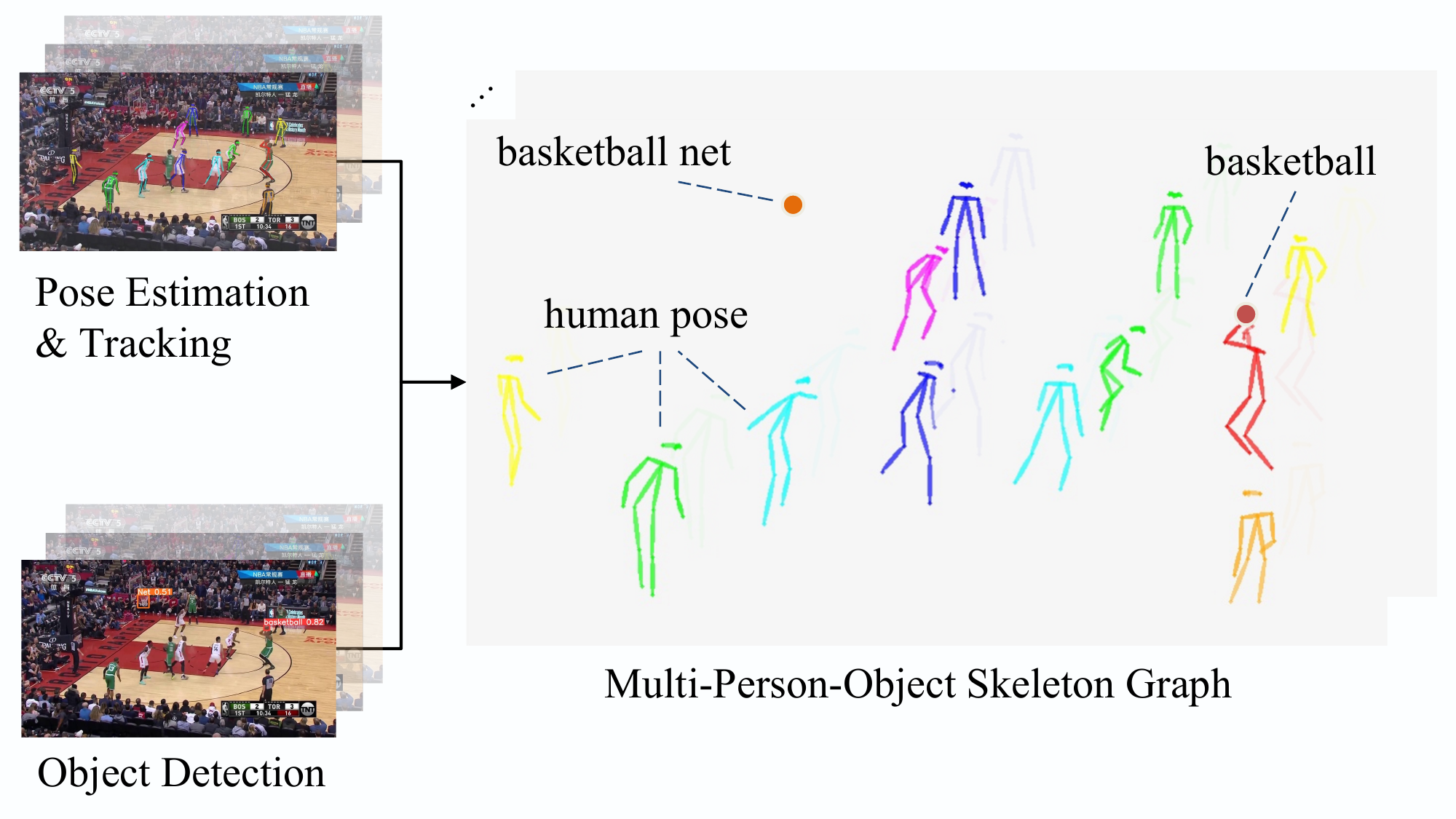}
    
    \caption{A group activity is represented as a panoramic graph which consists of multi-person skeletons and object keypoints.}
    \label{fig:idea}
\end{figure}

    To overcome the shortcomings of RGB-based methods, several recent approaches~\cite{gavrilyuk2020actortransformers, li2021groupformer, perez2022skeletonbased, zhou2022composer, thilakarathne2022pose} have attempted to leverage human pose for group activity recognition. These approaches extract keypoints using pose estimation backbones and take keypoints either as the sole input~\cite{perez2022skeletonbased, zhou2022composer, thilakarathne2022pose} or as a supplement to RGB inputs~\cite{gavrilyuk2020actortransformers, li2021groupformer}. However, these methods lack effective structural prior for the relationships of keypoints, thus requiring relational reasoning among them~\cite{perez2022skeletonbased}, or specialized hierarchical models to incrementally aggregate from keypoints to group-level features~\cite{zhou2022composer}. Furthermore, they rely on ground-truth individual bounding boxes and ball annotations for accurate localization and tracking before recognition.

    Considering joints as nodes and bones as edges, the human pose can naturally construct a skeletal graph. This topology provides a strong prior, enhancing the feature extraction capabilities of spatial and temporal graph convolutions~\cite{yan2018stgcn}. Consequently, Graph Convolutional Networks (GCNs) have achieved great success in skeleton-based action recognition~\cite{yan2018stgcn,shi2019-2sagcn,song2020resgcn,chen2021ctrgcn}. However, existing approaches are primarily based on the single-person skeletal topology. When multiple participants are involved, these approaches typically divide the multi-person skeletal data into multiple single-person graphs and stack them in the sample dimension. Under this processing, the weights of multiple people are shared within the network, resulting in the interactions between individuals being largely neglected~\cite{zhu2021drgcn, li2023twoperson}. Moreover, the absence of object data in conventional skeleton datasets~\cite{liu2020ntu} leads to a significant loss of critical information related to objects.
    


    To address the aforementioned shortcomings, we introduce a novel pipeline for skeleton-based GAR. This pipeline begins with the estimation of human poses from videos and the construction of a multi-person skeleton graph, followed by the recognition using a spatial-temporal graph convolution network. Initially, we employ pre-trained pose estimation and object detection methods~\cite{fang2023alphapose, sun2019hrnet, jocher2023yolov8} to extract keypoints for both humans and objects. Instead of relying on ground-truth tracklets, we develop a reassignment strategy coupled with pose tracking algorithms~\cite{aharon2022botsort,wang2020jde} to enhance consistent identity assignments across frames. Subsequently, as illustrated in~\cref{fig:idea}, we design a panoramic multi-person-object graph that integrates the keypoints of multiple human skeletons and objects. This panoramic graph overcomes the limitations of vanilla human skeleton representations, including the constrained single-person graph scale and the lack of object information. Lastly, we propose MP-GCN for skeleton-based GAR.
    Extensive experimental evaluations on Volleyball, NBA, and Kinetics demonstrate that the proposed method outperforms state-of-the-art keypoint-based methods. The contributions are summarized as follows:
    
    \begin{itemize}
        \item We develop a new pipeline for skeleton-based group activity recognition that does not require ground-truth individual boxes and labels. It includes acquiring keypoints through pose estimation and tracking algorithms and recognizing activities using skeleton-based graph convolutional network.
        
        \item We design a panoramic multi-person-object graph to represent group activity, which addresses the shortcomings of previous methods and unifies intra-person and inter-person interaction modeling.
        
        \item Using only estimated human pose and object keypoints, our method outperforms previous approaches on three widely used datasets with significantly lower computation cost compared to RGB-based methods. 
    \end{itemize}

    
\section{Related Work}
\label{sec:related}

\subsection{Group Activity Recognition}

    GAR~\cite{choi2009what} has been extensively studied due to its wide-ranging applications in the real world. The majority of the existing methods rely on RGB video either as the sole input modality or as the primary modality combined with optical flow and pose data. 
    Early methods extracted hand-crafted features and utilized probabilistic graphical models~\cite{ryoo2011stochastic, lan2012discriminative, lan2012social, choi2012unified, wang2013bilinear, amer2014hirf, amer2016sum} or AND-OR graphs~\cite{amer2012costsensitive,amer2013monte,shu2015joint} to infer group activities. RNN-based methods~\cite{ibrahim2016volleyball, deng2016structure, bagautdinov2017social, li2017sbgar, shu2017cern, wang2017recurrent, yan2018participationcontributed, ibrahim2018hierarchical, qi2018stagnet} were proposed due to RNN's capability for temporal modeling. Hierarchical LSTM models~\cite{ibrahim2016volleyball, wang2017recurrent, shu2017cern} were also proposed to capture spatial-temporal dependencies.

    Interaction modeling has attracted considerable attention in GAR. Researchers commonly employ interaction graphs~\cite{deng2016structure, qi2018stagnet, azar2019convolutional, wu2019learning, ehsanpour2020joint, hu2020progressive, yan2023higcin, yuan2021spatiotemporal} to depict the relationships between actors. Typically, they extract features using a CNN backbone with individual bounding boxes and learn interactions from the relation graph through RNNs~\cite{deng2016structure, qi2018stagnet}, GCNs~\cite{wu2019learning}, GATs~\cite{ehsanpour2020joint}, or more customized modules such as a dynamic inference network~\cite{yuan2021spatiotemporal} or a cross inference block~\cite{yan2023higcin}. Recently, transformer-based methods~\cite{gavrilyuk2020actortransformers, li2021groupformer, pramono2020empowering, pramono2021relational, kim2022detectorfree, zhou2022composer, pei2023key, zhu2023mlstformer} are predominant in GAR. These methods exploit attention mechanisms to model spatio-temporal relationship between individuals~\cite{han2022dualai, pei2023key, pramono2020empowering} or sub-groups~\cite{li2021groupformer, zhou2022composer}, integrating local information into group-level tokens for activity classification. 

    \noindent
    \textbf{Skeleton-based GAR.} Video-based methods for GAR often require substantial computational resources and encounter issues such as background variations and camera settings. As a result, several studies~\cite{perez2022skeletonbased, duan2022revisiting, zhou2022composer, thilakarathne2022pose, duan2023skeletr} explore using pose as the only input. Perez~\etal~\cite{perez2022skeletonbased} propose a model for interaction reasoning between human skeleton joints and ball keypoints. 
    Duan~\etal~\cite{duan2022revisiting} generate heatmap volumes using estimated 2D pose data from videos and feed them into 3D CNN for action recognition. 
    Zhou~\etal~\cite{zhou2022composer} develop a multi-scale transformer that hierarchically accumulates group tokens from original joint information.
    Another study by Duan~\etal~\cite{duan2023skeletr} captures short individual skeleton sequence using graph convolutions and models inter-sequence interactions using Transformer encoders in general action recognition tasks.
    In contrast to the aforementioned work, we introduce a multi-person-object graph to represent group activities and employ GCN to perform global graph convolution directly.

\subsection{GCN for Skeleton-based Action Recognition}

    Spatial Temporal GCNs~\cite{yan2018stgcn,shi2019-2sagcn,song2020resgcn,liu2020disentangling,chen2021ctrgcn} are widely adopted for modeling skeleton sequences in action recognition. They represent the movements of an individual as a spatial-temporal graph and perform the spatial-temporal graph convolution following~\cite{kipf2017semisupervised, yan2018stgcn}. However, these models face scalability issues when applied to multi-person scenarios. Conventional GCN methods split skeletons of multiple people, perform graph convolutions with shared weights, and average these single-person features to form a collective group representation for classification.

    \noindent
    \textbf{Skeleton-based Human Interaction Recognition.} Several GCN-based methods have attempted to model the interactive relations for recognition~\cite{zhu2021drgcn,gao2022aigcn,li2023twoperson}. Zhu~\etal~\cite{zhu2021drgcn} adopt inter-body graph convolution with a dynamic relational adjacency matrix, executed alongside the single-person graph convolution. 
    Li~\etal ~\cite{li2023twoperson} address the scalability issues of conventional GCN by applying graph convolution to a two-person graph that simultaneously encompasses inter-body and intra-body joint relationships. Inspired by these innovations, our method constructs a multi-person graph that depicts the interactions among all participants and objects in group activities.

    \noindent
    \textbf{Skeleton Representation with Object.} Some work integrates object information to enhance skeleton-based action recognition~\cite{xu2023skeletonbased,kim2019skeletonbased}. In contrast, our study focuses on introducing keypoint information in group activities. 
    
    To address the limitations in video and conventional skeleton representation, we propose a multi-person-object graph, which provides a panoramic view to depict all the individuals and objects involved in a group activity sequence.


\section{Method}
\label{sec:method}

\subsection{Preliminaries}

    \textbf{Human Skeleton Representation.} The skeleton representation uses $N$ joints and several bones to denote a person's body. In each frame, one body can be represented as a spatial graph
    $\mathcal{G}=(\mathcal{V}, \mathcal{E}, \mathcal{X})$,
    where
    $\mathcal{V}=\{v_{i}|i=1,...,N\}$
    is the vertex set of $N$ joints, and $\mathcal{E}=\{e_{ij}|i,j=1,...N\}$ is the edge set. An undirected edge $e_{ij}\in\mathcal{E}$
    represents the bone connection between $v_i$ and $v_j$. The relational adjacency matrix $\mathbf{A}\in\mathbb{R}^{N\times N}$ denotes all edges with its element $a_{ij}$ representing the strength of $e_{ij}$. 
    $\mathcal{X}\in\mathbb{R}^{T\times N\times C}$
    is the $C$-channel feature map of a sequence in $T$ frames, typically the 2D/3D joint coordinates. 

    \noindent
    \textbf{Graph Convolution.} Most GCN-based methods follow a similar paradigm. The backbone of these methods typically consists of several basic blocks. Each basic block is composed of a Spatial Graph Convolution (SGC) layer and a Temporal Convolution Network (TCN) layer, which captures the spatial relationships among skeleton joints within a single frame and the temporal dependencies across multiple frames, respectively. Yan~\etal~\cite{yan2018stgcn} describe the SGC operation as 
    \begin{equation}
        \label{eq:sgc1}
        f_{\text {out }}\left(v_{i}\right)=\sum_{v_{j} \in B\left(v_{ i}\right)} \frac{1}{Z_{i}\left(v_{j}\right)} f_{in}\left(v_{ j}\right) \cdot \mathbf{w}\left(l_{i}\left(v_{j}\right)\right),
    \end{equation}
    where $f_{in}()$ and $f_{out}()$ are input and output features of the corresponding joints, respectively. The node $v_{i}$ denotes the $i$-th joint, with $B(v_{i})$ representing its neighbor set. The labeling function $l_{i}:B(v_{i})\to \{0,...,K-1\}$ assigns each neighbor node to one of $K$ subsets. That weight function $\mathbf{w}$ is applied to the categorized neighbor nodes, and $Z_{i}(v_{j})$ serves as a normalizing term to balance the influence of different subsets.


\begin{figure*}[ht]
  \centering
  \includegraphics[width=\textwidth]{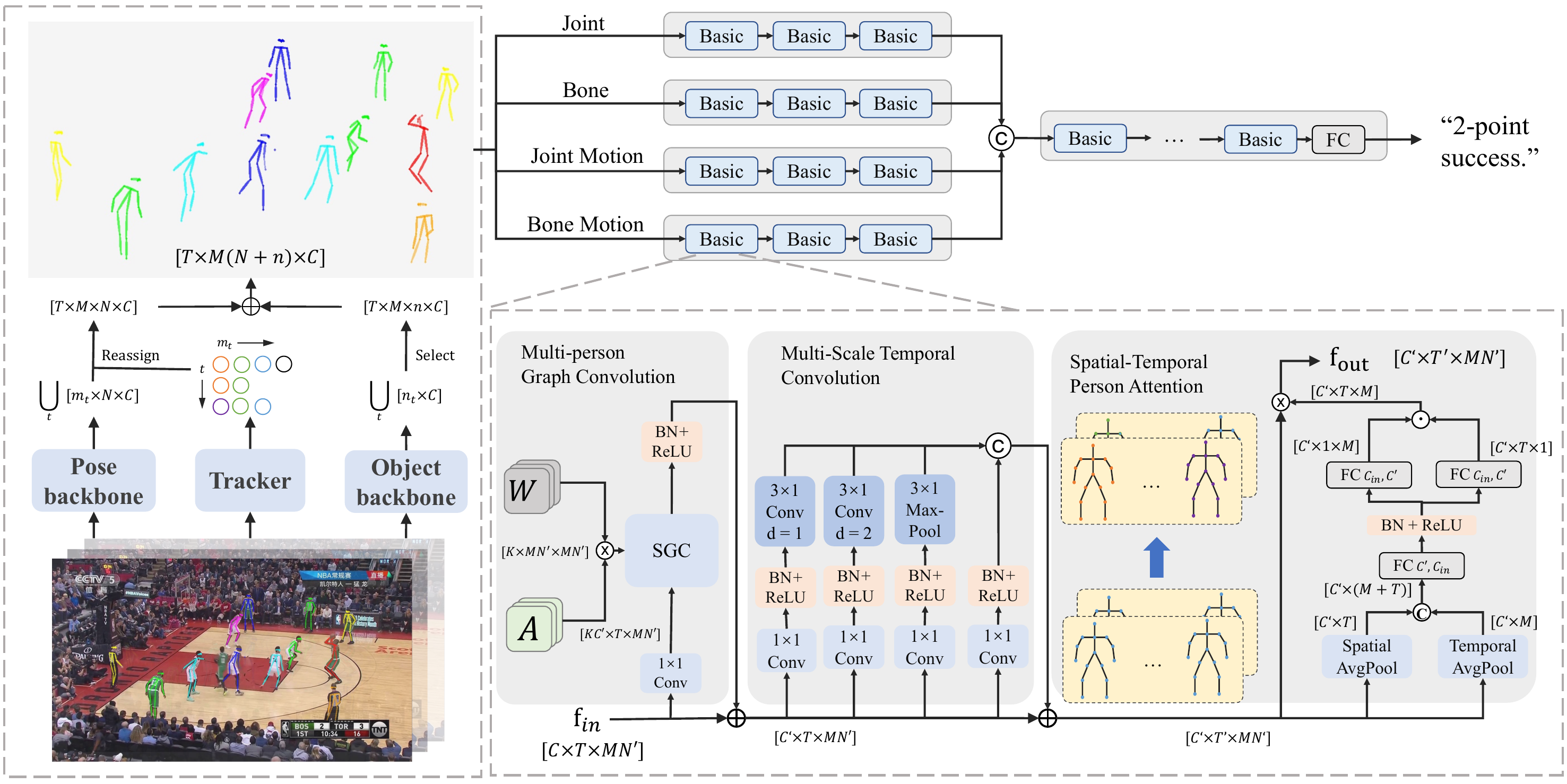}
  \caption{Architecture of MP-GCN and components of the basic block, where $C,C',T,T',N'$ and $K$ denote the numbers of input channels, output channels, input frames, output frames, joints, and subsets in SGC, respectively. $\odot$, $\otimes$, and © represent the matrix production, element-wise production, and concatenation, respectively.}
  \label{fig:network}
\end{figure*}

\subsection{Panoramic Graph Convolution}
    \noindent
    \textbf{Human-Object Graph.} For scenes involving multiple people, an activity sequence is represented by $\mathcal{X}\in\mathbb{R}^{T\times M\times N\times C}$, where $M$ is the number of participants. This skeleton representation only includes the coordinates of the human body, thus it often suffers from the absence of object information~\cite{liu2020ntu}. To address this issue, we introduce keypoints to represent objects and integrate them with human pose to form a human-object graph. The skeletal sequence is expanded to $\mathcal{X}\in\mathbb{R}^{T\times M\times N'\times C}$, where $N'=(N+n)$. Notably, the number of objects keypoints $n$ and how they connect with human skeletons can be arbitrarily defined. In this paper, we introduce the ball as one point corresponding to the center of its bounding box and empirically connect it with both hands.

    \noindent
    \textbf{Multi-Person Graph.} In contrast to prior work~\cite{yan2018stgcn, shi2019-2sagcn, liu2020disentangling, chen2021ctrgcn} that process multiple skeletons by separating them and batching them in the sample domain, our approach views all individuals in a scene as a unified group. Given an activity involving $M$ participants, we organize the activity sequence as a multi-person graph $\mathcal{X}\in\mathbb{R}^{T\times MN\times C}$ with an adjacency matrix $\mathbf{A}\in\mathbb{R}^{MN\times MN}$. $\mathbf{A}$ is partitioned into blocks $\mathbf{A}_{ij}\in\mathbb{R}^{N\times N}$. Specifically,
    \begin{equation}
        \mathbf{A} = 
        \begin{bmatrix}
            \mathbf{A}_{11} & \mathbf{A}_{12} & \cdots & \mathbf{A}_{1M} \\
            \mathbf{A}_{21} & \mathbf{A}_{22} & \cdots & \mathbf{A}_{2M}\\
            \vdots & \vdots & \ddots & \vdots\\
            \mathbf{A}_{M1} & \mathbf{A}_{M2} & \cdots & \mathbf{A}_{MM}
        \end{bmatrix}
    \end{equation}
    where the diagonal blocks $\mathbf{A}_{ii}$ capture the intra-body connections for each individual $i$, while the off-diagonal blocks represent inter-body relationships. 
    To construct this graph, we define the natural body topology as the intra-body graph and introduce pairwise inter-body links. These links connect the central body joints and object keypoints for every pair of individuals.
    Afterwards, we create a panoramic multi-person-object graph by integrating the concepts of human-object and multi-person graphs. This graph consists of spatial-temporal features $\mathcal{X}\in\mathbb{R}^{T\times MN'\times C}$ with an expanded adjacency matrix $\mathbf{A}\in\mathbb{R}^{MN'\times MN'}$.
    Compared to scene graphs that reduce each identity to a single node or the conventional single-person skeletons, the proposed panoramic graph provides a more detailed and holistic representation of the group activity.

    \noindent
    \textbf{Spatial Graph Convolution.} The graph convolution operation is applied to extract information embedded in the multi-person-object skeleton graph. When implemented with adjacency matrix $\mathbf{A}$, \cref{eq:sgc1} can be articulated as
    \begin{equation}
    \label{eq:sgc2}
        \mathbf{f}_{out}=\sum_{k=0}^{K-1} \mathbf{\Lambda}_{k}^{-\frac{1}{2}} \mathbf{A}_{k} \mathbf{\Lambda}_{k}^{-\frac{1}{2}} \mathbf{f}_{in} \mathbf{W}_{k}
    \end{equation}
    In this paper, we maintain the maximum graph sampling distance $D$ to 1 and set the number of subsets $K$ to 3 as in ~\cite{yan2018stgcn}. We design an intra-inter partitioning strategy to accommodate multi-person graph structure. In \cref{eq:sgc2}, $\mathbf{f}_{in}$ and $\mathbf{f}_{out}$ denote the input and output feature maps. The matrices $\mathbf{A}_0$, $\mathbf{A}_1$ and $\mathbf{A}_2$ correspond to self links, intra-person connections and inter-person connections, respectively. $\mathbf{\Lambda}_k$ is used to normalize $\mathbf{A}_k$, and the learnable parameter $\mathbf{W}_k$ represents the edge importance weighting. This partitioning strategy allows GCN to capture both intra-person and inter-person interactions through a singular graph convolution framework.

\subsection{Model Architecture}

    Building on the multi-person spatial graph convolution, we propose MP-GCN for skeleton-based GAR, as illustrated in \cref{fig:network}.
    Current state-of-the-art methods~\cite{li2021groupformer, gavrilyuk2020actortransformers, shi2019-2sagcn,chen2021ctrgcn} often employ a multi-stream architecture and fuse the prediction scores at the final stage. Such ensemble techniques like late-fusion are beneficial for improving model performance, but they linearly increase model complexity with the addition of input modalities. Instead, we incorporate ideas from~\cite{song2020resgcn, liu2020disentangling, gavrilyuk2020actortransformers} to construct an early-fusion network. Our MP-GCN contains four input branches and a main branch. Each input branch contains several basic blocks to extract low-level features from one of four aspects of skeletal data: joint, bone, joint motion, and bone motion (detailed preprocessing steps are provided in the Appendix). Subsequently, we concatenate the feature maps from the four streams into the main branch at an early stage. The main branch continues with additional basic blocks and classification head.
    This architecture retains rich information from different inputs but significantly reduces complexity compared to conventional ensemble solutions.

    \noindent
    \textbf{Basic Block.} 
    The basic block consists of an SGC, Multi-Scale TCN, and Spatial-Temporal Person Attention block with module-wise residual links. In the SGC, graph convolution is performed on the entire panoramic graph, maintaining the graph scale throughout the network. For the TCN, we propose a four-branch multi-scale temporal convolution layer following~\cite{liu2020disentangling} to capture temporal dynamics in consecutive frames. Each branch contains a bottleneck layer for channel transformation, BatchNorm and ReLU, and ends with $3\times1$ convolutional layers with specified dilation in the first two branches and a Max-Pooling layer in the third branch. 

    \noindent
    \textbf{Spatial-Temporal Person Attention.} 
    A group activity is often characterized by key roles that can vary rapidly over time~\cite{pei2023key}. To capture the key individuals throughout an activity, we design a Spatial-Temporal Person Attention Module, based on \cite{song2023effgcn}. It computes attention scores for $M$ people across $T$ frames. It starts by averaging the feature maps in person and frame dimensions. Then, it concatenates the averaged features into a $C\times(M+T)$ vector. Subsequently, two convolutional layers are designed to derive person-wise and frame-wise scores, which are multiplied to generate the final spatial-temporal attention map. This attention module enables MP-GCN to focus on key individuals and objects in crowded scenes and to mitigate the impact of irrelevant roles.

\subsection{Tracking-based Pose Reassignment} 

    Despite the widespread application of motion capture and ball tracking systems in sports analysis~\cite{2023nba}, there are few publicly available datasets for skeleton-based GAR. Therefore, we extract 2D skeleton coordinates from original RGB videos using pre-trained pose estimators and object detectors~\cite{fang2023alphapose, sun2019hrnet, jocher2023yolov8}. 

    Human Pose Tracking have proven to be vital for action recognition~\cite{rajasegaran2023benefits, duan2023skeletr}, yet their adoption for associating skeleton data in the respective of recognition remains limited. For a group activity sequence, the extracted human pose can be denoted as $ \bigcup_{t}\mathcal{X'}_t$, where $\mathcal{X'}_t\in\mathbb{R}^{m_t\times N\times C}$ and the number of detected actors $m_t$ may change along $t$. 
    We leverage a ReID tracker that assigns unique IDs to the same person identified across frames. However, occlusions, miss detections and the identification of non-relevant individuals occur frequently. 
    To enhance data consistency and filter out irrelevant individuals when preparing the skeletal data, we develop a pose assignment strategy which maps a variable number of individuals per frame to a fixed number $M$ using $r: N^+ \to \{0,...,M-1\}$.
    

    \noindent
    \textbf{Reassignment strategy.} 
    Assuming $m_t$ people are detected in frame $t$, each person is assigned a track ID $id(m)$ and an associated detection confidence $conf(m)$. For the individual with index $m$, we define the sequence $S\_id(m)={(x_t, y_t)}_{t=1}^{t_{id(m)}}$ which represents the trajectory of person $m$'s center coordinates $(x_t, y_t)$ over $t_{id(m)}$ frames identified by the same track ID $id(m)$. The activeness of a person with track ID $e(id(m))$ is evaluated using $e(id(m)) =\exp({s(id(m))})/\Sigma_{i=1}^{m_t}\exp({s(id(i))})$, where
    \begin{equation}
    \label{eq:energy}
        s(id(m)) = \sqrt{\sum_{t=1}^{t_{id(m)}}\frac{(x_t - \bar{x})^2}{t_{id(m)}}}+\sqrt{\sum_{t=1}^{t_{id(m)}}\frac{(y_t - \bar{y})^2}{t_{id(m)}}}
    \end{equation}
    In \cref{eq:energy}, $\bar{x}=1/t_{id(m)}\sum_{t=1}^{t_{id(m)}}x_t$ and $\bar{y}=1/t_{id(m)}\sum_{t=1}^{t_{id(m)}}y_t$. $s(id(m))$ signifies the standard deviation of the trajectory $S\_id(m)$. The pose reassignment contains two stages. First, in each frame, it selects individuals with at most highest $M$ scores, where the score is determined by $score(id(m))=conf(m)+e(id(m))$. We map the chosen track IDs to a fixed range using $r(id(m)) = id(m)\mod{M}$ for the first assignment. In case of a conflict, the person with smaller track ID will be assigned, while the others are reserved for the second assignment. Subsequently, the remaining individuals are allocated to the available indices in order. This reassignment strategy prioritizes who are more likely to play a significant role in group activities and enhance temporal consistency of the obtained skeletal data. 

\section{Experiments}

\subsection{Datasets}
\noindent
\textbf{Volleyball Dataset}~\cite{ibrahim2016volleyball} contains 4,830 clips from 55 volleyball game records. Each clip, consisting of 41 frames, is labeled with one of eight group activities. The authors also provide labels of individual actions (9 classes) and bounding boxes for every athlete in the central frame. For fully supervised studies, researchers commonly use the group activities, individual action labels, and the ground-truth person bounding box of the middle 20 frames~\cite{sendo2019heatmapping}. We report the Multi-class Classification Accuracy (MCA) for group activity and the Mean Classification Accuracy of individual action (IMCA) on \underline{Volleyball Fully Supervised} following most methods. In contrast, weakly supervised methods only utilize group-level activity labels. On \underline{Volleyball Weakly Supervised}, we follow previous methods~\cite{yan2020nba, kim2022detectorfree} and report MCA and Merged-MCA (MMCA, which merges \textit{left/right set} and \textit{left/right pass} into \textit{left/right pass-net}).

\noindent
\textbf{NBA Dataset.} NBA~\cite{yan2020nba} contains 9,172 clips (7,624 for training and 1,548 for testing) derived from 181 NBA game records. Each clip, composed of 72 frames, is labeled with one of nine basketball scoring activities. This is the largest dataset currently for group activity recognition, providing only the group-level activity annotations. For evaluation, we report the Multi-Class Accuracy (MCA) and Mean Per-Class Accuracy (MPCA), consistent with previous methods.


\subsection{Implementation Details}

    \noindent
    \textbf{Keypoint Acquisition.} We adopt YOLOv8~\cite{jocher2023yolov8} for pose estimation and object detection. We utilize YOLOv8x-pose with weights pre-trained on the COCO dataset ~\cite{lin2014microsoft} for pose estimation ($V$=17), and employ BoTSort~\cite{aharon2022botsort} for multi-person tracking. For both the Volleyball and NBA datasets, we set the number of actors in each frame $M$ to 12. Additionally, we fine-tune the pretrained YOLOv8x model on our custom dataset for detecting basketballs and basketball nets on the NBA dataset. This dataset includes 5,000 sample images manually annotated by us from the NBA dataset. Please refer to Appendix for details about the keypoint extraction process.

    \noindent
    \textbf{Hyper Parameters.} We set the number of frames $T$ to the original length of each clip, $T=41$ for Volleyball Weakly Supervised and $T=72$ for NBA dataset. In four input branches, the input-output channels for 3 basic blocks are 6-64, 64-64, 64-32. In the main branch, the channels for the 6 basic blocks are 128-128, 128-128, 128-128, 128-256, 256-256, 256-256. The training and testing batch size for both the Volleyball and NBA datasets is set to 16. We train our model for 65 epochs, employing a warm-up strategy~\cite{he2016deep} for the first 5 epochs to ensure stable training. The learning rate is initially set to 0.1 and decays according to a cosine scheduler after the 5th epoch. We use cross-entropy loss as the training objective and an SGD optimizer with Nesterov momentum of 0.9 and weight decay of 0.0002. 

\subsection{Comparison with State of the Arts}


\begin{table}[t]
\begin{center}
\caption{Comparison with state-of-the-art methods on the Volleyball dataset. 
The best results are highlight in \textbf{bold}, and the second-best results are \underline{underlined}.} 
\label{tab:volleyball}
\begin{tabular}{lcccccc}
\hline
Method    & Keypoint & RGB & Flow & Backbone & \multicolumn{2}{c}{Benchmark} \\ 
\hline
\addlinespace[0.5ex]
\multicolumn{5}{l}{\textbf{Fully Supervised}} & MCA & IMCA \\
\addlinespace[1ex]
ARG~\cite{wu2019learning}                & & \checkmark & & VGG-19 & 92.6 & 82.6 \\ %
HiGCIN~\cite{yan2023higcin}              & & \checkmark & & ResNet-18 & 91.4 & - \\
DIN~\cite{yuan2021spatiotemporal}        & & \checkmark & & VGG-16 & 93.6 & - \\ %
GIRN~\cite{perez2022skeletonbased}       & \checkmark & & & OpenPose & 92.2 & - \\
POGARS~\cite{thilakarathne2022pose}      & \checkmark & & & Hourglass & 93.9 & - \\
COMPOSER~\cite{zhou2022composer}         & \checkmark & & & HRNet & 94.6 & - \\
SkeleTR~\cite{duan2023skeletr}           & \checkmark & & & HRNet & 94.4 & - \\
\rowcolor{lightgray}MP-GCN (Ours)        & \checkmark & & & HRNet & 95.5 & 84.6 \\
CRM~\cite{azar2019convolutional}         & & \checkmark & \checkmark & I3D & 93.0 & - \\
AT~\cite{gavrilyuk2020actortransformers} & \checkmark & \checkmark & & I3D+HRNet & 93.5 & 85.7 \\ %
SACRF~\cite{pramono2020empowering}       & \checkmark & \checkmark & \checkmark & I3D+AlphaPose & 95.0 & 83.1 \\ %
GroupFormer~\cite{li2021groupformer}     & \checkmark & \checkmark & \checkmark & I3D+AlphaPose & \underline{95.7} & 85.6 \\
Dual-AI~\cite{han2022dualai}             & & \checkmark & \checkmark & Inception-v3 & 95.4 & 85.3 \\

\rowcolor{lightgray}MP-GCN (Ours) + VGG16  & \checkmark & \checkmark & & HRNet+VGG16 & \textbf{96.2} & - \\
\addlinespace[1ex]\hline
\addlinespace[0.5ex]
\multicolumn{5}{l}{\textbf{Weakly Supervised}} & MCA & MMCA \\
\addlinespace[1ex]
SAM~\cite{yan2020nba}
    & & \checkmark & & ResNet-18 & 86.3  & 93.1 \\
DFWSGAR~\cite{kim2022detectorfree}
    & & \checkmark & & ResNet-18 & 90.5  & 94.4 \\
Dual-AI~\cite{han2022dualai}
    & & \checkmark & \checkmark & Inception-v3 & - & \underline{95.8} \\
KRGFormer~\cite{pei2023key}
    & & \checkmark & & Inception-v3 & 92.4  & 95.0 \\
Tamura\etal~\cite{tamura2022hunting}
    & & \checkmark & & I3D+DETR &\textbf{96.0} & - \\
\rowcolor{lightgray}MP-GCN (Ours)
    & \checkmark & & & YOLOV8x & \underline{92.8} & \textbf{96.1} \\
\hline
\end{tabular}
\end{center}
\end{table}

\noindent
\textbf{Volleyball Dataset.} The results are presented in \cref{tab:volleyball}. For a fair comparison, results for fully supervised settings (using groundtruth box) and for weakly supervised settings (using estimated detections) are listed separately. In the fully supervised setting, we utilize the skeletal data extracted by COMPOSER~\cite{zhou2022composer}, which employs HRNet~\cite{sun2019hrnet} with GT human bbox~\cite{sendo2019heatmapping} ($T$=20). We report the results for previous RGB-based or multi-modal methods~\cite{wu2019learning, azar2019convolutional, yan2023higcin, gavrilyuk2020actortransformers, pramono2020empowering, yuan2021spatiotemporal, li2021groupformer, han2022dualai} and results for keypoint-based methods~\cite{gavrilyuk2020actortransformers, perez2022skeletonbased, zhou2022composer} that rely solely on human pose and ball annotations~\cite{perez2022skeletonbased}. With strong supervision provided by ground-truth individual bounding boxes for human pose tracking, our method achieves a remarkable result: 95.5\% in MCA, outperforming both keypoint-based and RGB-based approaches. Meanwhile, our method can perform individual action recognition using single-person skeleton graph, obtaining 84.6\% in IMCA. Moreover, we combine our model's prediction with scores of VGG-16 backbone (which achieves 92.3\% on Volleyball) and finally obtain 96.2\% in MCA. This result demonstrates our method's effectiveness in predicting the group activity from a complex scene and its further complementarity with RGB-based methods. In the weakly supervised setting, ground-truth bounding boxes are replaced with estimated tracklets obtained using YOLOv8~\cite{jocher2023yolov8, aharon2022botsort}. As shown in ~\cref{tab:volleyball}, our method outperforms others under both fully and weakly supervised settings. In the weakly supervised setting, our method still achieves satisfying results: 92.8\% MCA and 96.1\% MMCA. This result demonstrate our method's compatibility to both high-quality and noisy keypoints data.


\begin{table}[t]
    \centering
    \caption{Comparison with state-of-the-art methods on the NBA dataset. $^*$ denotes that MCA and MPCA are reported in~\cite{kim2022detectorfree}.}
    \label{table:nba}
    \begin{tabular}{lc|@{\hspace{0.5em}}c@{\hspace{0.3em}}c@{\hspace{1em}}cc}
    \toprule
        Method & Input & \#Params & FLOPs & MCA & MPCA \\ 
    \midrule
        $^*$ARG~\cite{wu2019learning} 
        & RGB & 45.4M & 614.2G & 59.0 & 56.8\\
        $^*$AT~\cite{gavrilyuk2020actortransformers}
        & RGB & 29.6M & 609.9G & 47.1 & 41.5 \\
        SAM~\cite{yan2020nba}
        & RGB & - & - & 49.1 & 47.5 \\
        $^*$DIN~\cite{yuan2021spatiotemporal}
        & RGB & 28.6M & 613.8G & 61.6 & 56.0 \\
        dual-AI\cite{han2022dualai} 
        & RGB & - & - & 51.5 & 44.8 \\
        KRGFormer~\cite{pei2023key} 
        & RGB & - & - & 72.4 & 67.1 \\
        DFWSGAR\cite{kim2022detectorfree}
        & RGB & 17.4M & 628.1G & 75.8 & 71.2 \\
    \midrule
        MP-GCN (Ours, $T$=18)
        & kpt & 4.4M & 5.70G & 73.3 & 68.4 \\
        MP-GCN (Ours, $T$=72)
        & kpt & 4.4M & 22.3G & \underline{76.0}  & \underline{71.9} \\
        MP-GCN (Ours, late-fusion) 
        & kpt & 11.6M & 50.4G & \textbf{78.7} & \textbf{74.6} \\
    \bottomrule
    \end{tabular}
\end{table}

\noindent
\textbf{NBA Dataset.} \Cref{table:nba} presents the state-of-the-art comparison on the NBA dataset. 
The CNN backbone for RGB-based methods is set to ResNet-18 with an input resolution of 720×1280 and $T$=18 (except for \cite{han2022dualai}, which uses Inception-v3 and $T$=3/20 for training/testing). We follow the same pose estimation implementation as in Volleyball Weakly Supervised and also detect balls and basketball nets using YOLOV8~\cite{jocher2023yolov8}. Furthermore, for methods with available official implementation, we also report the model's number of parameters and FLOPs in the recognition phase. Note that \#Params and FLOPs of the backbone for RGB-based methods are included since the CNN backbone is tuned during training, whereas the computational cost of our pose backbone is only considered for skeleton dataset acquisition and is not included. Our method outperforms video-based methods in MCA with significantly fewer parameters and lower computational costs. When late-fusion is adopted, our method significantly surpasses video-based methods. Please refer to Appendix for a detailed comparison between keypoint-based methods~\cite{yan2018stgcn, liu2020disentangling, chen2021ctrgcn, lee2023hdgcn} in terms of accuracy, computational cost and backbone efficiency.

\subsection{Ablation Study}

\noindent
\textbf{Graph Structure.} Suppose $P$ denotes the number of graphs for each sequence and $N$ denotes the number of points in a graph. To validate the proposed multi-person graph, we experiment with the following graph settings on the Volleyball dataset. (1) Baseline: each sample is represented by $M$ weight-shared graphs of $V$ nodes. (2) Baseline with non-shared weights: $M$ single-person graphs with non-shared weights. (3) Panoramic intra-person graph: a multi-person graph with only intra-person edges. (4) Panoramic graph: a multi-person-object graph with both intra- and inter-person connections. As presented in \cref{tab:graph}, using the multi-person graph scale results in a significant improvement of about 1.5\% with tolerable increases in parameters and FLOPs. These results demonstrate the effectiveness the panoramic graph structure. Additionally, \Cref{tab:M} presents an ablation study of $M$ on the NBA dataset and demonstrate that $M=12$ obtains the highest accuracy. This finding is consistent with the typical composition of a basketball game, which involves 10 athletes and 2-3 referees. It also suggests that the proposed reassignment strategy effectively maintains the key roles on the court.

\begin{table}[t]
    \centering
    \caption{Comparison of different graph structures in number of parameters, FLOPs and MCA (\%) on Volleyball Fully Supervised.}
    \label{tab:graph}
    \begin{tabular}{lcc|ccc}
    \toprule
        Graph Structure &  $P$  & $N$ & \#Param. & FLOPs & MCA \\ 
    \midrule
        Baseline                        & 12 & $V$   & 1.47M & 2.21G & 93.66 \\ 
        Baseline (non-shared weights)   & 12 & $V$   & 1.78M & 2.21G & 94.54 \\
        Panoramic (only intra links)    & 1  & 12$V$ & 3.70M & 4.19G & 95.21 \\
        Panoramic (intra+inter links)   & 1  & 12$V$ & 3.70M & 4.19G & \textbf{95.54} \\
    \bottomrule
    \end{tabular}
\end{table}

\begin{table}[t]
    \begin{minipage}[t]{0.28\textwidth}
        \caption{Comparison of different $M$ in MCA (\%) on the NBA dataset.}
        \label{tab:M}
        \centering
        \begin{tabular}{l|c}
        \toprule
            $M$ & MCA \\
        \midrule
            3  & 71.71\\
            6  & 73.45\\
            9  & 74.61 \\
            \textbf{12 (Ours)} & \textbf{75.96} \\ 
            15 & 74.49 \\
        \bottomrule
        \end{tabular}
    \end{minipage}
    \hfill
    \begin{minipage}[t]{0.7\textwidth}
        \caption{Comparison of different graph structures where objects are organized differently in MCA (\%) on NBA dataset.}
        \label{tab:graph2}
        \centering
        \begin{tabular}{lc|c}
        \toprule
            Graph Structure &  $N$  & MCA \\ 
        \midrule
            pose                     & $V$ & 65.83 \\
            pose*$M$                  & 12$V$ & 69.21 \\ 
            pose*$M$ + ball             & 12$V$+1 & 69.51 \\
            pose*$M$ + ball + net        & 12$V$+2 & 73.06 \\ 
            (pose-ball)*$M$            & 12($V$+1) & 72.94 \\
            (pose-ball-net)*$M$(Ours) & 12($V$+2) & \textbf{75.96} \\
        \bottomrule
        \end{tabular}
    \end{minipage}
\end{table}

\noindent
\textbf{Effect of Object.} To further investigate the effect of object information, we examined graph structures incorporating objects in various configurations, as presented in \cref{tab:graph2}. \underline{pose*$M$+\{ball/+net\}} consists of $MV+v$ nodes, including $M$ people and $v$ object points (ball/+basketball net) connected to the hands of each actor. \underline{(pose-\{ball/+net\})*$M$} introduces individual object nodes for each person, which are initialized with the same data. Placing the object keypoints separately from the multi-person skeletons leads to significantly lower accuracy compared to having individual non-shared object keypoints. Additionally, the inclusion of extra objects, such as the basketball net, further enhances recognition accuracy, indicating the effectiveness and generalizability of the proposed panoramic graph.


\subsection{Visualization}

\begin{figure*}[t]
    \centering
    \includegraphics[width=\linewidth]{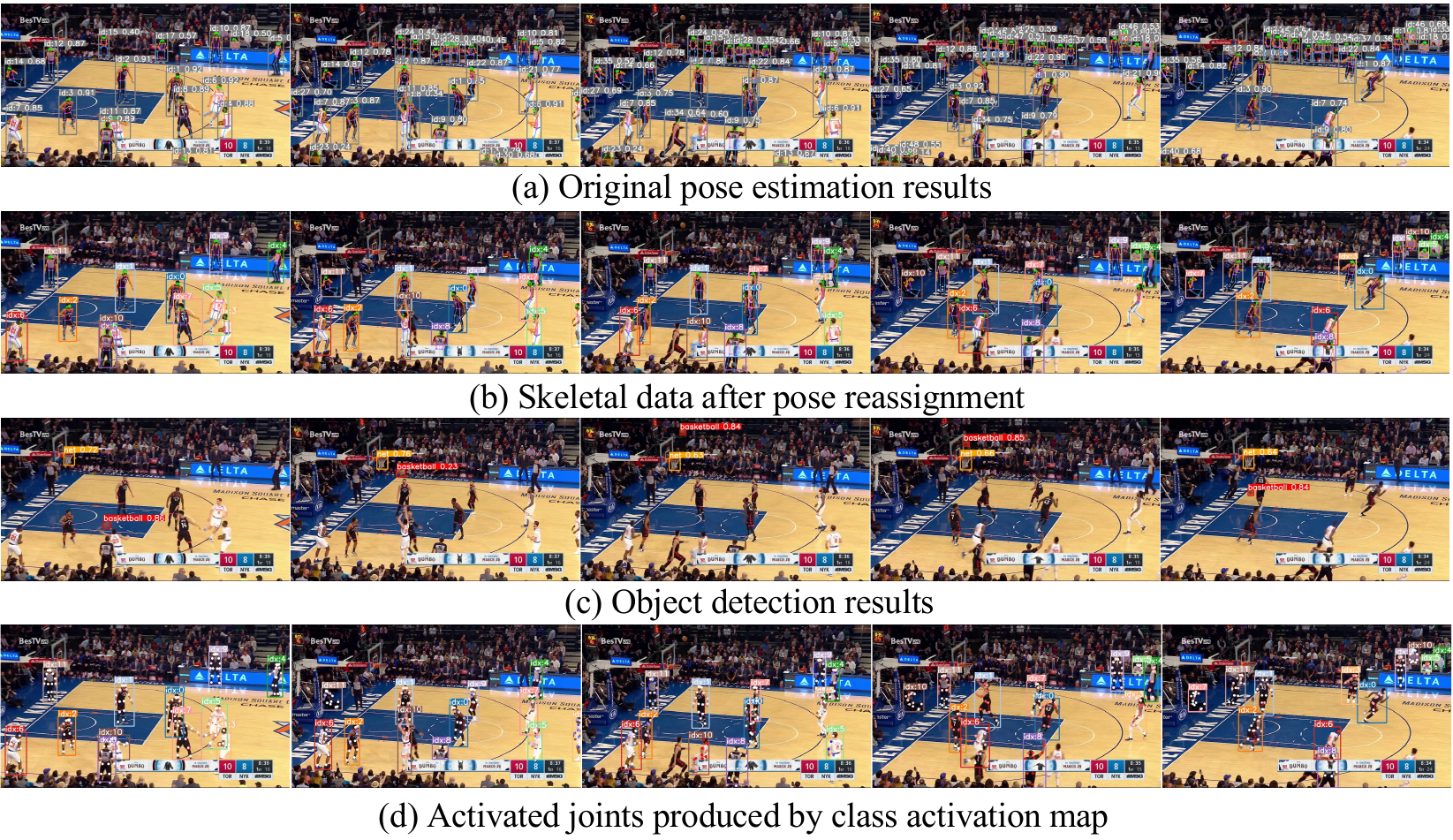}
    \caption{Visualization of the extracted skeletons, object keypoints and activation maps.}
    \label{fig:vis}
\end{figure*}

\begin{figure*}[t]
    \centering
    \includegraphics[width=0.88\linewidth]{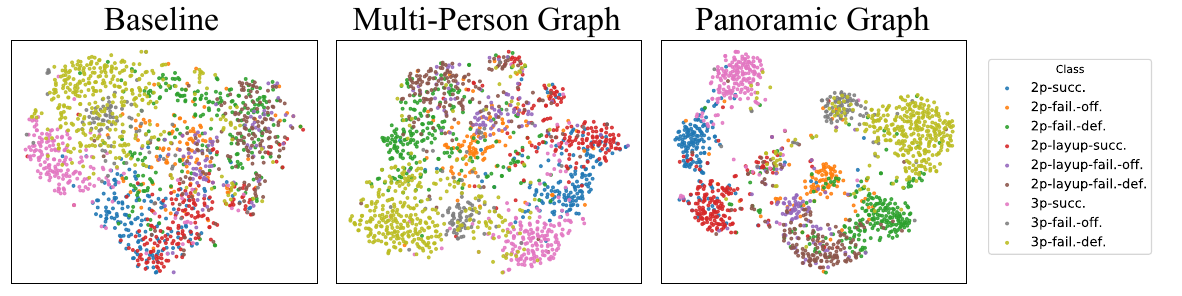}
    \caption{$t$-SNE feature embedding visualization on NBA dataset for different graphs.}
    \label{fig:tsne}
\end{figure*}

\noindent
\textbf{Visualization of Keypoints and Activation Maps.} In \cref{fig:vis}, we visualize the human skeletons and object keypoints for a 5-frame clip from the NBA dataset. In \cref{fig:vis} (a)-(b), it is evident that the proposed reassignment strategy effectively selects athletes and referees from a multitude of detected candidates. However, when the number of people in the scene is insufficient, it inevitably includes unrelated individuals off the court. 
In \cref{fig:vis} (c), our method effectively detects fast-moving basketballs. In addition, we visualize the class activation map~\cite{zhou2016learning} of this sequence in \cref{fig:vis} (d). Our model correctly focuses on the shooter and referee when the shooter attempts a two-point shot. 

\noindent
\textbf{Visualization of Embedding Feature.} \Cref{fig:tsne} presents the $t$-SNE~\cite{van2008visualizing} visualization of different graph structures. It is evident that the proposed multi-person-object graph facilitates a clear separation of each class, while also producing a reasonable distribution of similar classes. For instance, the "2-point-layup-failed-offensive-rebound" and "2-point-layup-failed-defensive-rebound" samples are closely located in the embedding space.

\subsection{Towards in-the-wild Skeleton-based Action Recognition}

    To validate the generalization of our method, we also tested it on the Kinetics400 dataset~\cite{kay2017kinetics}, which has over 300k videos across 400 action classes, presenting significant challenges for skeleton-based action recognition. 
    We use the same skeletal data provided by Duan \etal~\cite{duan2022revisiting} and observe that 90\% of the samples contain fewer than 4 individuals on average. Thus, we set $M$ to 4 and keep other parameters unchanged. \Cref{tab:k400} presents the comparison between skeleton-based methods. 
    It's noticed that our method also achieves the remarkable accuracy on this daily living dataset. Notably, the introduction of object keypoints by Hachiuma\etal~\cite{hachiuma2023unified} substantially enhances recognition accuracy. This finding underscores the value of object information and suggests that the performance could be significantly improved by integrating object keypoints.


\begin{table}[tb]
    \centering
    \caption{SOTA Comparison of keypoint-based methods on Kinetics400 dataset.}
    \label{tab:k400}
    \begin{tabular}{lcc}
    \toprule
    Method                & Pose Estimator & Acc. \\
    \midrule
    ST-GCN~\cite{yan2018stgcn} & \multirow{3}{*}{OpenPose~\cite{cao2021openpose}} 
        & 30.7 \\
    AGCN~\cite{shi2019-2sagcn} &  
        & 36.1 \\
    MS-G3D~\cite{liu2020disentangling} &  
        & 38.0 \\
    \midrule
    Hachiuma~\etal~\cite{hachiuma2023unified} & \multirow{2}{*}{PPNv2~\cite{sekii2018pose}}
        & 43.1 \\
    Hachiuma~\etal with objects~\cite{hachiuma2023unified} &
        & \textbf{52.3} \\
    \midrule
    MS-G3D~\cite{liu2020disentangling} & \multirow{3}{*}{HRNet~\cite{sun2019hrnet}}
        & 45.1 \\
    PoseConv3D(\textit{\textbf{J}}+\textit{\textbf{B}})~\cite{duan2022revisiting} &  
        & 47.7 \\
    Hachiuma~\etal~\cite{hachiuma2023unified}  &  
        & 50.3 \\
    \midrule
    MP-GCN(Ours) & \multirow{2}*{HRNet~\cite{sun2019hrnet}}
        & 48.4 \\
    MP-GCN(Ours, late-fusion) & 
        & \underline{51.1} \\
    \bottomrule
    \end{tabular}
\end{table}
    


    
    While our method has shown promising results, it is important to acknowledge that the performance of skeleton-based methods fall short of SOTA video-based methods~\cite{piergiovanni2023rethinking}, which obtain over 90\% of accuracy on Kinetics400. 
    This gap highlights the potential for future research to focus on enhancing multi-person and object representations, particularly in uncontrolled environments.
    


\section{Conclusion}

This paper proposes a new pipeline for Skeleton-based Group Activity Recognition. This pipeline includes human and object keypoints estimation, tracking-based pose reassignment, and collective activity recognition through panoramic graph convolutional network. The panoramic graph incorporates human pose and object keypoints into a multi-person-object skeleton graph, addressing the limitations of traditional human skeleton representation in shared weight, lack of inter-person interaction modeling, and absence of object information. Using skeletal data as the only input, the proposed method achieves state-of-the-art performance on three widely-used datasets (Volleyball, NBA, and Kinetics).
\section*{Acknowledgements}

This work was supported by National Natural Science Foundation of China (grant No. 62376068, grant No. 62350710797), by Guangdong Basic and Applied Basic Research Foundation (grant No. 2023B1515120065),  by Guangdong S\&T programme (grant No. 2023A0505050109), by Shenzhen Science and Technology Innovation Program (grant No. JCYJ20220818102414031).

\bibliographystyle{splncs04}
\bibliography{abrv_review, main}

\begin{thebibliography}{10}
\providecommand{\url}[1]{\texttt{#1}}
\providecommand{\urlprefix}{URL }
\providecommand{\doi}[1]{https://doi.org/#1}

\bibitem{aharon2022botsort}
Aharon, N., Orfaig, R., Bobrovsky, B.Z.: Bot-sort: Robust associations
  multi-pedestrian tracking. arXiv:2206.14651  (2022)

\bibitem{amer2016sum}
Amer, M.R., Todorovic, S.: Sum product networks for activity recognition. IEEE
  TPAMI  \textbf{38}(4),  800--813 (2016)

\bibitem{amer2013monte}
Amer, M.R., Todorovic, S., Fern, A., Zhu, S.C.: Monte carlo tree search for
  scheduling activity recognition. In: ICCV. pp. 1353--1360 (2013)

\bibitem{amer2012costsensitive}
Amer, M.R., Xie, D., Zhao, M., Todorovic, S., Zhu, S.C.: Cost-sensitive
  top-down/bottom-up inference for multiscale activity recognition. In: ECCV.
  pp. 187--200 (2012)

\bibitem{amer2014hirf}
Amer, M.R., Lei, P., Todorovic, S.: Hirf: Hierarchical random field for
  collective activity recognition in videos. In: ECCV. pp. 572--585 (2014)

\bibitem{azar2019convolutional}
Azar, S.M., Atigh, M.G., Nickabadi, A., Alahi, A.: Convolutional relational
  machine for group activity recognition. In: CVPR. pp. 7884--7893 (2019)

\bibitem{bagautdinov2017social}
Bagautdinov, T., Alahi, A., Fleuret, F., Fua, P., Savarese, S.: Social scene
  understanding: End-to-end multi-person action localization and collective
  activity recognition. In: CVPR. pp. 3425--3434 (2017)

\bibitem{cao2021openpose}
Cao, Z., Hidalgo, G., Simon, T., Wei, S.E., Sheikh, Y.: Openpose: Realtime
  multi-person 2d pose estimation using part affinity fields. IEEE TPAMI
  \textbf{43}(1),  172--186 (2021)

\bibitem{chen2021ctrgcn}
Chen, Y., Zhang, Z., Yuan, C., Li, B., Deng, Y., Hu, W.: Channel-wise topology
  refinement graph convolution for skeleton-based action recognition. In: ICCV.
  pp. 13339--13348 (2021)

\bibitem{choi2012unified}
Choi, W., Savarese, S.: A unified framework for multi-target tracking and
  collective activity recognition. In: ECCV. pp. 215--230 (2012)

\bibitem{choi2009what}
Choi, W., Shahid, K., Savarese, S.: What are they doing? : Collective activity
  classification using spatio-temporal relationship among people. In: ICCVW.
  pp. 1282--1289 (2009)

\bibitem{deng2016structure}
Deng, Z., Vahdat, A., Hu, H., Mori, G.: Structure inference machines: Recurrent
  neural networks for analyzing relations in group activity recognition. In:
  CVPR. pp. 4772--4781 (2016)

\bibitem{duan2023skeletr}
Duan, H., Xu, M., Shuai, B., Modolo, D., Tu, Z., Tighe, J., Bergamo, A.:
  Skeletr: Towards skeleton-based action recognition in the wild. In: ICCV. pp.
  13588--13598 (2023)

\bibitem{duan2022revisiting}
Duan, H., Zhao, Y., Chen, K., Lin, D., Dai, B.: Revisiting skeleton-based
  action recognition. In: CVPR. pp. 2959--2968 (2022)

\bibitem{ehsanpour2020joint}
Ehsanpour, M., Abedin, A., Saleh, F., Shi, J., Reid, I., Rezatofighi, H.: Joint
  learning of social groups, individuals action and sub-group activities in
  videos. In: ECCV. pp. 177--195 (2020)

\bibitem{fang2023alphapose}
Fang, H.S., Li, J., Tang, H., Xu, C., Zhu, H., Xiu, Y., Li, Y.L., Lu, C.:
  Alphapose: Whole-body regional multi-person pose estimation and tracking in
  real-time. IEEE TPAMI  \textbf{45}(6),  7157--7173 (2023)

\bibitem{gao2022aigcn}
Gao, F., Xia, H., Tang, Z.: Attention interactive graph convolutional network
  for skeleton-based human interaction recognition. In: ICME. pp.~1--6 (2022)

\bibitem{gavrilyuk2020actortransformers}
Gavrilyuk, K., Sanford, R., Javan, M., Snoek, C.G.M.: Actor-transformers for
  group activity recognition. In: CVPR. pp. 836--845 (2020)

\bibitem{hachiuma2023unified}
Hachiuma, R., Sato, F., Sekii, T.: Unified keypoint-based action recognition
  framework via structured keypoint pooling. In: CVPR. pp. 22962--22971 (2023)

\bibitem{han2022dualai}
Han, M., Zhang, D.J., Wang, Y., Yan, R., Yao, L., Chang, X., Qiao, Y.: Dual-ai:
  Dual-path actor interaction learning for group activity recognition. In:
  CVPR. pp. 2990--2999 (2022)

\bibitem{2023nba}
hawkeyeinnovations: Nba and sony's hawk-eye innovations launch strategic
  partnership powering next generation tracking technology.
  https://pr.nba.com/nba-sony-hawk-eye-innovations-partnership/ (2023)

\bibitem{he2016deep}
He, K., Zhang, X., Ren, S., Sun, J.: Deep residual learning for image
  recognition. In: CVPR. pp. 770--778 (2016)

\bibitem{hu2020progressive}
Hu, G., Cui, B., He, Y., Yu, S.: Progressive relation learning for group
  activity recognition. In: CVPR. pp. 977--986 (2020)

\bibitem{ibrahim2018hierarchical}
Ibrahim, M., Mori, G.: Hierarchical relational networks for group activity
  recognition and retrieval. In: ECCV. pp. 721--736 (2018)

\bibitem{ibrahim2016volleyball}
Ibrahim, M.S., Muralidharan, S., Deng, Z., Vahdat, A., Mori, G.: A hierarchical
  deep temporal model for group activity recognition. In: CVPR. pp. 1971--1980
  (2016)

\bibitem{jocher2023yolov8}
Jocher, G., Chaurasia, A., Qiu, J.: Yolo by ultralytics (2023)

\bibitem{kay2017kinetics}
Kay, W., Carreira, J., Simonyan, K., Zhang, B., Hillier, C., Vijayanarasimhan,
  S., Viola, F., Green, T., Back, T., Natsev, P., Suleyman, M., Zisserman, A.:
  The kinetics human action video dataset. arXiv:1705.06950  (2017)

\bibitem{kim2022detectorfree}
Kim, D., Lee, J., Cho, M., Kwak, S.: Detector-free weakly supervised group
  activity recognition. In: CVPR (2022)

\bibitem{kim2019skeletonbased}
Kim, S., Yun, K., Park, J., Choi, J.Y.: Skeleton-based action recognition of
  people handling objects. In: WACV. pp. 61--70 (2019)

\bibitem{kipf2017semisupervised}
Kipf, T.N., Welling, M.: Semi-supervised classification with graph
  convolutional networks. arXiv:1609.02907  (2017)

\bibitem{lan2012social}
Lan, T., Sigal, L., Mori, G.: Social roles in hierarchical models for human
  activity recognition. In: CVPR. pp. 1354--1361 (2012)

\bibitem{lan2012discriminative}
Lan, T., Wang, Y., Yang, W., Robinovitch, S.N., Mori, G.: Discriminative latent
  models for recognizing contextual group activities. IEEE TPAMI
  \textbf{34}(8),  1549--1562 (2012)

\bibitem{lee2023hdgcn}
Lee, J., Lee, M., Lee, D., Lee, S.: Hierarchically decomposed graph
  convolutional networks for skeleton-based action recognition. In: ICCV. pp.
  10444--10453 (2023)

\bibitem{li2021groupformer}
Li, S., Cao, Q., Liu, L., Yang, K., Liu, S., Hou, J., Yi, S.: Groupformer:
  Group activity recognition with clustered spatial-temporal transformer. In:
  ICCV. pp. 13648--13657 (2021)

\bibitem{li2017sbgar}
Li, X., Chuah, M.C.: Sbgar: Semantics based group activity recognition. In:
  ICCV. pp. 2895--2904 (2017)

\bibitem{li2023twoperson}
Li, Z., Li, Y., Tang, L., Zhang, T., Su, J.: Two-person graph convolutional
  network for skeleton-based human interaction recognition. IEEE TCSVT
  \textbf{33}(7),  3333--3342 (2023)

\bibitem{lin2014microsoft}
Lin, T.Y., Maire, M., Belongie, S., Hays, J., Perona, P., Ramanan, D., Dollar,
  P., Zitnick, C.L.: Microsoft coco: Common objects in context. In: ECCV. pp.
  740--755 (2014)

\bibitem{liu2020ntu}
Liu, J., Shahroudy, A., Perez, M., Wang, G., Duan, L.Y., Kot, A.C.: Ntu rgb+d
  120: A large-scale benchmark for 3d human activity understanding. IEEE TPAMI
  \textbf{42}(10),  2684--2701 (2020)

\bibitem{liu2020disentangling}
Liu, Z., Zhang, H., Chen, Z., Wang, Z., Ouyang, W.: Disentangling and unifying
  graph convolutions for skeleton-based action recognition. In: CVPR. pp.
  140--149 (2020)

\bibitem{pei2023key}
Pei, D., Huang, D., Kong, L., Wang, Y.: Key role guided transformer for group
  activity recognition. IEEE TCSVT  \textbf{33}(12),  7803--7818 (2023)

\bibitem{perez2022skeletonbased}
Perez, M., Liu, J., Kot, A.C.: Skeleton-based relational reasoning for group
  activity analysis. PR  \textbf{122},  108360 (2022)

\bibitem{piergiovanni2023rethinking}
Piergiovanni, A.J., Kuo, W., Angelova, A.: Rethinking video vits: Sparse video
  tubes for joint image and video learning. In: CVPR. pp. 2214--2224 (2023)

\bibitem{pramono2020empowering}
Pramono, R.R.A., Chen, Y.T., Fang, W.H.: Empowering relational network by
  self-attention augmented conditional random fields for group activity
  recognition. In: ECCV. pp. 71--90 (2020)

\bibitem{pramono2021relational}
Pramono, R.R.A., Fang, W.H., Chen, Y.T.: Relational reasoning for group
  activity recognition via self-attention augmented conditional random field.
  IEEE TIP  \textbf{30},  8184--8199 (2021)

\bibitem{qi2018stagnet}
Qi, M., Qin, J., Li, A., Wang, Y., Luo, J., Van~Gool, L.: stagnet: An attentive
  semantic rnn for group activity recognition. In: ECCV. pp. 104--120 (2018)

\bibitem{rajasegaran2023benefits}
Rajasegaran, J., Pavlakos, G., Kanazawa, A., Feichtenhofer, C., Malik, J.: On
  the benefits of 3d pose and tracking for human action recognition. In: CVPR
  (2023)

\bibitem{ryoo2011stochastic}
Ryoo, M.S., Aggarwal, J.K.: Stochastic representation and recognition of
  high-level group activities. IJCV  \textbf{93}(2),  183--200 (2011)

\bibitem{sekii2018pose}
Sekii, T.: Pose proposal networks. In: ECCV. pp. 342--357 (2018)

\bibitem{sendo2019heatmapping}
Sendo, K., Ukita, N.: Heatmapping of people involved in group activities. In:
  MVA. pp.~1--6 (2019)

\bibitem{shi2019-2sagcn}
Shi, L., Zhang, Y., Cheng, J., Lu, H.: Two-stream adaptive graph convolutional
  networks for skeleton-based action recognition. In: CVPR. pp. 12018--12027
  (2019)

\bibitem{shu2017cern}
Shu, T., Todorovic, S., Zhu, S.C.: Cern: Confidence-energy recurrent network
  for group activity recognition. In: CVPR. pp. 4255--4263 (2017)

\bibitem{shu2015joint}
Shu, T., Xie, D., Rothrock, B., Todorovic, S., Chun~Zhu, S.: Joint inference of
  groups, events and human roles in aerial videos. In: CVPR. pp. 4576--4584
  (2015)

\bibitem{song2020resgcn}
Song, Y.F., Zhang, Z., Shan, C., Wang, L.: Stronger, faster and more
  explainable: A graph convolutional baseline for skeleton-based action
  recognition. pp. 1625--1633 (2020)

\bibitem{song2023effgcn}
Song, Y.F., Zhang, Z., Shan, C., Wang, L.: Constructing stronger and faster
  baselines for skeleton-based action recognition. IEEE TPAMI  \textbf{45}(2),
  1474--1488 (2023)

\bibitem{sun2019hrnet}
Sun, K., Xiao, B., Liu, D., Wang, J.: Deep high-resolution representation
  learning for human pose estimation. In: CVPR. pp. 5686--5696 (2019)

\bibitem{sun2023surveyHARmultimodal}
Sun, Z., Ke, Q., Rahmani, H., Bennamoun, M., Wang, G., Liu, J.: Human action
  recognition from various data modalities: A review. IEEE TPAMI
  \textbf{45}(3),  3200--3225 (2023)

\bibitem{tamura2022hunting}
Tamura, M., Vishwakarma, R., Vennelakanti, R.: Hunting group clues with
  transformers for social group activity recognition. In: ECCV. pp. 19--35
  (2022)

\bibitem{thilakarathne2022pose}
Thilakarathne, H., Nibali, A., He, Z., Morgan, S.: Pose is all you need: The
  pose only group activity recognition system (pogars). MVA  \textbf{33}(6),
  ~95 (2022)

\bibitem{van2008visualizing}
{Van der Maaten}, L., Hinton, G.: Visualizing data using t-sne. JMLR
  \textbf{9}(11) (2008)

\bibitem{wang2017recurrent}
Wang, M., Ni, B., Yang, X.: Recurrent modeling of interaction context for
  collective activity recognition. In: CVPR. pp. 7408--7416 (2017)

\bibitem{wang2013bilinear}
Wang, Z., Shi, Q., Shen, C., {van den Hengel}, A.: Bilinear programming for
  human activity recognition with unknown mrf graphs. In: CVPR. pp. 1690--1697
  (2013)

\bibitem{wang2020jde}
Wang, Z., Zheng, L., Liu, Y., Li, Y., Wang, S.: Towards real-time multi-object
  tracking. In: ECCV. pp. 107--122 (2020)

\bibitem{wu2019learning}
Wu, J., Wang, L., Wang, L., Guo, J., Wu, G.: Learning actor relation graphs for
  group activity recognition. In: CVPR. pp. 9956--9966 (2019)

\bibitem{xu2023skeletonbased}
Xu, L., Lan, C., Zeng, W., Lu, C.: Skeleton-based mutually assisted interacted
  object localization and human action recognition. IEEE TMM  \textbf{25},
  4415--4425 (2023)

\bibitem{yan2018participationcontributed}
Yan, R., Tang, J., Shu, X., Li, Z., Tian, Q.: Participation-contributed
  temporal dynamic model for group activity recognition. In: ACM MM. pp.
  1292--1300 (2018)

\bibitem{yan2020nba}
Yan, R., Xie, L., Tang, J., Shu, X., Tian, Q.: Social adaptive module for
  weakly-supervised group activity recognition. In: ECCV. pp. 208--224 (2020)

\bibitem{yan2023higcin}
Yan, R., Xie, L., Tang, J., Shu, X., Tian, Q.: Higcin: Hierarchical graph-based
  cross inference network for group activity recognition. IEEE TPAMI
  \textbf{45}(6),  6955--6968 (2023)

\bibitem{yan2018stgcn}
Yan, S., Xiong, Y., Lin, D.: Spatial temporal graph convolutional networks for
  skeleton-based action recognition. In: AAAI. pp. 7444--7452 (2018)

\bibitem{yuan2021learning}
Yuan, H., Ni, D.: Learning visual context for group activity recognition. In:
  AAAI. pp. 3261--3269 (2021)

\bibitem{yuan2021spatiotemporal}
Yuan, H., Ni, D., Wang, M.: Spatio-temporal dynamic inference network for group
  activity recognition. In: ICCV. pp. 7476--7485 (2021)

\bibitem{zhou2016learning}
Zhou, B., Khosla, A., Lapedriza, A., Oliva, A., Torralba, A.: Learning deep
  features for discriminative localization. In: CVPR. pp. 2921--2929 (2016)

\bibitem{zhou2022composer}
Zhou, H., Kadav, A., Shamsian, A., Geng, S., Lai, F., Zhao, L., Liu, T.,
  Kapadia, M., Graf, H.P.: Composer: Compositional reasoning of group activity
  in videos with keypoint-only modality. In: ECCV. pp. 249--266 (2022)

\bibitem{zhu2021drgcn}
Zhu, L., Wan, B., Li, C., Tian, G., Hou, Y., Yuan, K.: Dyadic relational graph
  convolutional networks for skeleton-based human interaction recognition. PR
  \textbf{115},  107920 (2021)

\bibitem{zhu2023mlstformer}
Zhu, X., Zhou, Y., Wang, D., Ouyang, W., Su, R.: Mlst-former: Multi-level
  spatial-temporal transformer for group activity recognition. IEEE TCSVT
  \textbf{33}(7),  3383--3397 (2023)

\end{thebibliography}



\end{document}